\documentclass[a4paper,twoside]{article}

\usepackage{threeparttable}
\usepackage{hyperref} 
\usepackage{tabu}
\usepackage{multirow}
\usepackage{xcolor}
\usepackage{url} 
\usepackage{booktabs}
\usepackage{multirow}
\usepackage{siunitx}
\usepackage{diagbox}
\usepackage{epsfig}
\usepackage{subcaption}
\usepackage{calc}
\usepackage{amssymb}
\usepackage{amstext}
\usepackage{amsmath}
\usepackage{amsthm}
\usepackage{multicol}
\usepackage{pslatex}
\usepackage{apalike}
\usepackage{algorithm2e}
\usepackage{multirow}
\usepackage[bottom]{footmisc}
\usepackage{booktabs}
\usepackage{tabularx}
\usepackage{SCITEPRESS}     

\begin{document}

\title{Patient Trajectory Prediction: Integrating Clinical Notes with Transformers
}

\author{\authorname{Sifal Klioui \sup{1},  Sana Sellami \sup{1} and Youssef Trardi \sup{1}}
\affiliation{\sup{1}Aix-Marseille Univ, LIS, CNRS Marseille, France}
\email{first\_author@etu.univ-amu.fr, second\_author@univ-amu.fr, third\_author@univ-amu.fr}
}


\abstract{Predicting disease trajectories from electronic health records (EHRs) is a complex task due to major challenges such as data non-stationarity, high granularity of medical codes, and integration of multimodal data. EHRs contain both structured data, such as diagnostic codes, and unstructured data, such as clinical notes, which hold essential information often overlooked. Current models, primarily based on structured data, struggle to capture the complete medical context of patients, resulting in a loss of valuable information.
To address this issue, we propose an approach that integrates unstructured clinical notes into transformer-based deep learning models for sequential disease prediction. This integration enriches the representation of patients' medical histories, thereby improving the accuracy of diagnosis predictions. Experiments on MIMIC-IV datasets demonstrate that the proposed approach outperforms traditional models relying solely on structured data.}

\onecolumn \maketitle \normalsize \setcounter{footnote}{0} \vfill

\section{\uppercase{Introduction}}
\label{sec:introduction}

\begin{figure*}[h]
\centering
\includegraphics[width=1\linewidth]{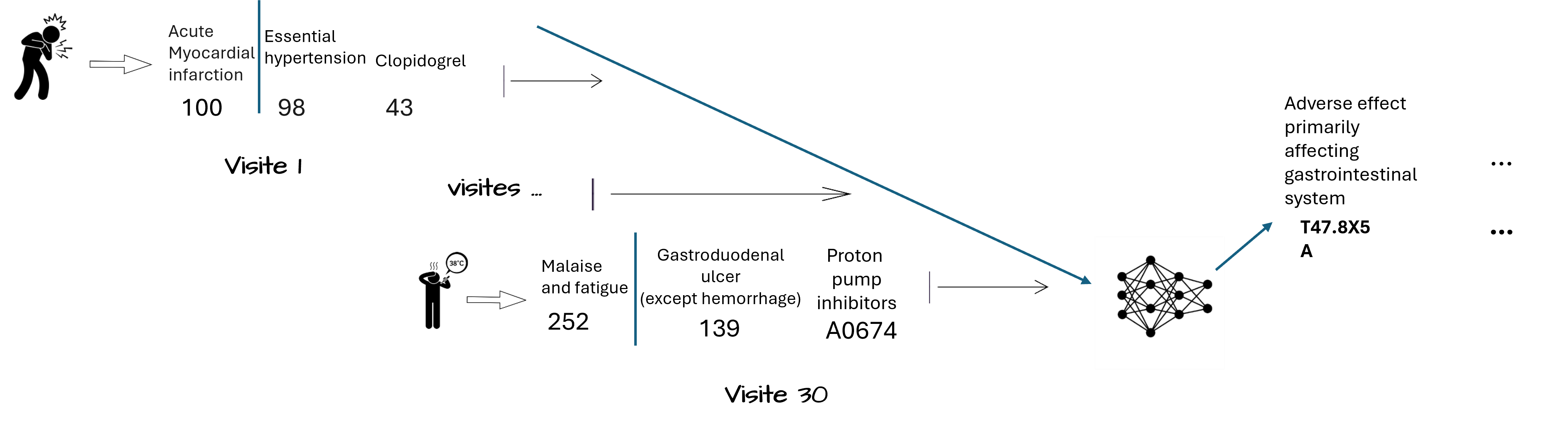}
\caption{Sequential disease predictions}
\label{fig:Prédiction séquentielle des maladies}
\end{figure*}

In healthcare, the exponential growth of Electronic Health Records (EHRs) has revolutionized patient care while posing new challenges. Healthcare professionals now frequently interact with medical records spanning several decades, having to process and analyze this vast amount of information to make informed decisions about patients' future health status. This evolution has accelerated the development of automated systems to predict future diagnoses from past medical data, thus becoming a key element of personalized and proactive medicine (Figure \ref{fig:Prédiction séquentielle des maladies}).
Machine learning techniques, particularly deep learning, have seen increasing growth in medicine \cite{egger2022medical}, thanks to their adaptability and good results. In medical imaging, for example, deep learning models have achieved a high level of performance in predicting medical diagnoses, sometimes comparable to or even surpassing that of human experts \cite{mall2023comprehensive}. These results have led researchers to apply similar techniques to the task of sequential disease prediction (\cite{choi2016doctor,rodrigues2021lig,shankar2023clinical}), where the goal is to predict a patient's diagnosis at their next visit (N+1) based on the content of their previous visits (N). However, modeling patient trajectories from EHR data presents unique challenges:
\begin{itemize}
    \item The non-stationarity of EHR data which leads to variations in the data, limiting the generalizability of models.
    
    \item The high granularity of medical codes (e.g., over 70,000 in the International Classification of Diseases, 10th Revision, Clinical Modification (ICD-10-CM \footnote{\url{https://www.cdc.gov/nchs/icd/icd-10-cm/index.html}})) which makes it difficult for prediction models to explore and use these codes.

    \item Long-term dependencies due to processing long data sequences represent a difficult task for traditional recurrent neural network (RNN) models.

    \item  The integration of multimodal data as EHR data includes both structured information, such as laboratory results, and unstructured information, such as clinical notes.

    \item The impact of external factors (e.g., lifestyle, environment) that can lead to variability and uncertainty in predictions.
\end{itemize}

\noindent Addressing these challenges is essential to develop both accurate and reliable patient trajectory prediction systems capable of assisting physicians in decision-making by providing comprehensive forecasts based on a patient's clinical history.

In light of these challenges, this article focuses on improving the accuracy of automated medical prognosis systems, particularly in predicting future diagnoses based on patients' historical medical records. Current coding systems, such as the International Classification of Diseases (ICD) \footnote{\url{https://www.who.int/standards/classifications/classification-of-diseases}}, often do not fully capture the richness of information contained in clinical notes, which can lead to a loss of valuable information for predicting patient trajectories. 
To overcome this problem, we propose an approach aimed at improving the accuracy of diagnostic code predictions by integrating clinical note embeddings into transformers, which typically rely solely on medical codes. This method incorporates a discriminating factor that reduces prediction errors by enriching the representation of embeddings. This also allows for the recovery of valuable information often lost in coding systems such as ICD. By incorporating additional context, our approach addresses challenges related to understanding the reasons behind medication prescriptions, procedures performed, and diagnoses made.

This article is organized as follows: Section \ref{sec:related_work_and_challenges} reviews the literature. Section \ref{sec:Methods} describes our approach, including the process of generating embeddings and their integration into transformers. In Section \ref{sec:results_and_discussion}, we present our experimental results. Finally, Section \ref{sec:Conclusion} concludes this article and presents future work.

\section{State of the Art}
\label{sec:related_work_and_challenges}

Various methods, whether based on deep learning or traditional approaches, have been explored to predict patient trajectories. Among them, \textit{Doctor AI} \cite{choi2016doctor}, a temporal model based on recurrent neural networks (RNN), developed and applied to longitudinal time-stamped EHR data. \textit{Doctor AI} predicts a patient's medical codes and estimates the time until the next visit. However, it is limited by a fixed window width, which proves inadequate, as a patient's future diagnosis may depend on medical conditions outside this window. \textit{LIG-Doctor} \cite{rodrigues2021lig}, an artificial neural network architecture designed to efficiently predict patient trajectories using minimal bidirectional recurrent networks \textit{MGRU}. \textit{MGRU} handle the granularity of ICD-9 codes, but suffer from the same limitations as \textit{Doctor AI}. In \cite{choi2016retain}, the authors propose RETAIN, an interpretable predictive model for healthcare using reverse time attention mechanism.  Two RNNs are trained in reverse time order to learn the importance of previous visits, offering improved interpretability. \textit{DeepCare} \cite{pham2017predicting} employs Long Short-Term Memory (LSTM) networks for predicting next visit diagnosis codes, intervention recommendations and future risk prediction. The Life Model (LM) Framework \cite{manashty2019life} proposed an efficient representation of temporal data in concise sequences for training RNN-based models, introducing Mean Tolerance Error (MTE) as both a loss function and metric. \textit{Deep Patient} \cite{miotto2016deep} introduced an unsupervised deep learning approach using Stack Denoising Autoencoders (SDA) to extract meaningful feature representations from EHR data, but does not consider temporal characteristics, which is limiting, as this notion of time is inherent to the trajectory of a patient. In parallel, more classical methods such as Markov chains \cite{severson2020personalized}, Bayesian networks \cite{longato2022time}, and Hawkes processes \cite{lima2023hawkes} have been explored, but suffer from computational complexity when faced with massive data. 

The introduction of transformers marked an advancement, with \textit{Clinical GAN} \cite{shankar2023clinical}, a Generative Adversarial Networks (GAN) method based on the Transformer architecture. In this approach, an encoder-decoder model serves as the generator, while an encoder-only Transformer acts as the critic. The goal was to address exposure bias \cite{arora2022exposure}, a general issue (i.e., not specific to the Transformer) that arises from the teacher forcing training strategy. However, the use of GANs is challenged by scalability issues, such as training instability, non-convergence, and mode collapse \cite{saad2024survey}.

Despite the development of various approaches for predicting medical codes, it is important to note that most proposed models have been trained on electronic health records (EHRs) containing only structured data on diagnoses and procedures, such as ICD and CCS codes. However, these data omit some essential contextual information, such as medical reasoning and patient-specific nuances, which can be captured through clinical notes.

Moreover, comparing results between different studies poses several challenges:

\begin{itemize}
    \item \textbf{Dataset Variation}: Studies utilize different datasets (e.g., MIMIC-III vs. MIMIC-IV), which encompass varying patient populations and time periods \cite{johnson2016mimic,johnson2020mimic}. This variation can lead to discrepancies in results, as one dataset may present more challenging diagnoses to predict than another due to differing distributions. Consequently, such discrepancies complicate the reliability of comparisons between studies and may impact the applicability of findings to clinical practice.
    \item \textbf{Test Set Size}: The size of the test set can significantly impact results. For instance, Shankar et al. \cite{shankar2023clinical} used a test set of only 5\% of their dataset (approximately 1700 visits), which may not adequately represent the diversity of the patients' profiles and complexity.
    \item \textbf{Lack of Standardization}: There's often a lack of transparency regarding specific training datasets and preprocessing steps. Additionally, inconsistencies in the implementation of evaluation metrics can lead to discrepancies in reported results.
    \item \textbf{Preprocessing Variations}: Different preprocessing steps, such as tokenization and data cleaning, can affect model performance and hinder direct comparisons \cite{edin2023automated}.
    \item \textbf{Code Mapping Inconsistencies}: Some approaches predict medical codes directly, while others map to ICD codes. Variations in mapping schemes, such as choosing to apply the Clinical Classification Software Refined (CCSR) or not, can lead to inconsistencies in final code representations (i.e., different target labels).
\end{itemize}

These challenges underscore the importance of careful consideration when comparing results across different studies in this field. To enhance comparability and reproducibility in research on patient trajectory prediction, it is crucial to standardize datasets, preprocessing methods, and evaluation metrics.

\section{Proposed Methodology}
\label{sec:Methods}
In this section, we describe our approach for predicting patient trajectories, which relies on the MIMIC-IV datasets \footnote{\url{https://physionet.org/content/mimiciv/2.1/}} \footnote{\url{https://physionet.org/content/mimic-iv-note/2.2/}}.
We detail our methodology, including data preprocessing, model architecture, and integration of clinical notes.
\subsection{Data Preprocessing}

Preprocessing of MIMIC-IV data includes several operations: 

\begin{itemize}
    \item Extraction of diagnoses, procedures, and medications.
    \item Selection of patients with at least two visits.
    \item Exclusion of patients without all three types of medical codes \cite{shankar2023clinical}.
    \item Use of CCSR (Clinical Classification Software Refined) to map ICD-10-CM diagnoses into clinically significant categories, balancing CCS (Clinical Classifications Software) categories of ICD-9-CM with the specificity of ICD-10-CM, as well as ICD-10-PCS procedures, leveraging the specificity and taxonomy of the ICD-10-PCS coding scheme.
    \item Removal of infrequent codes (threshold of 5). \cite{edin2023automated}.
    \item Temporal ordering of events to create sequential trajectories.
\end{itemize}

\begin{figure*}[h]
\centering
\includegraphics[width=0.8\linewidth]{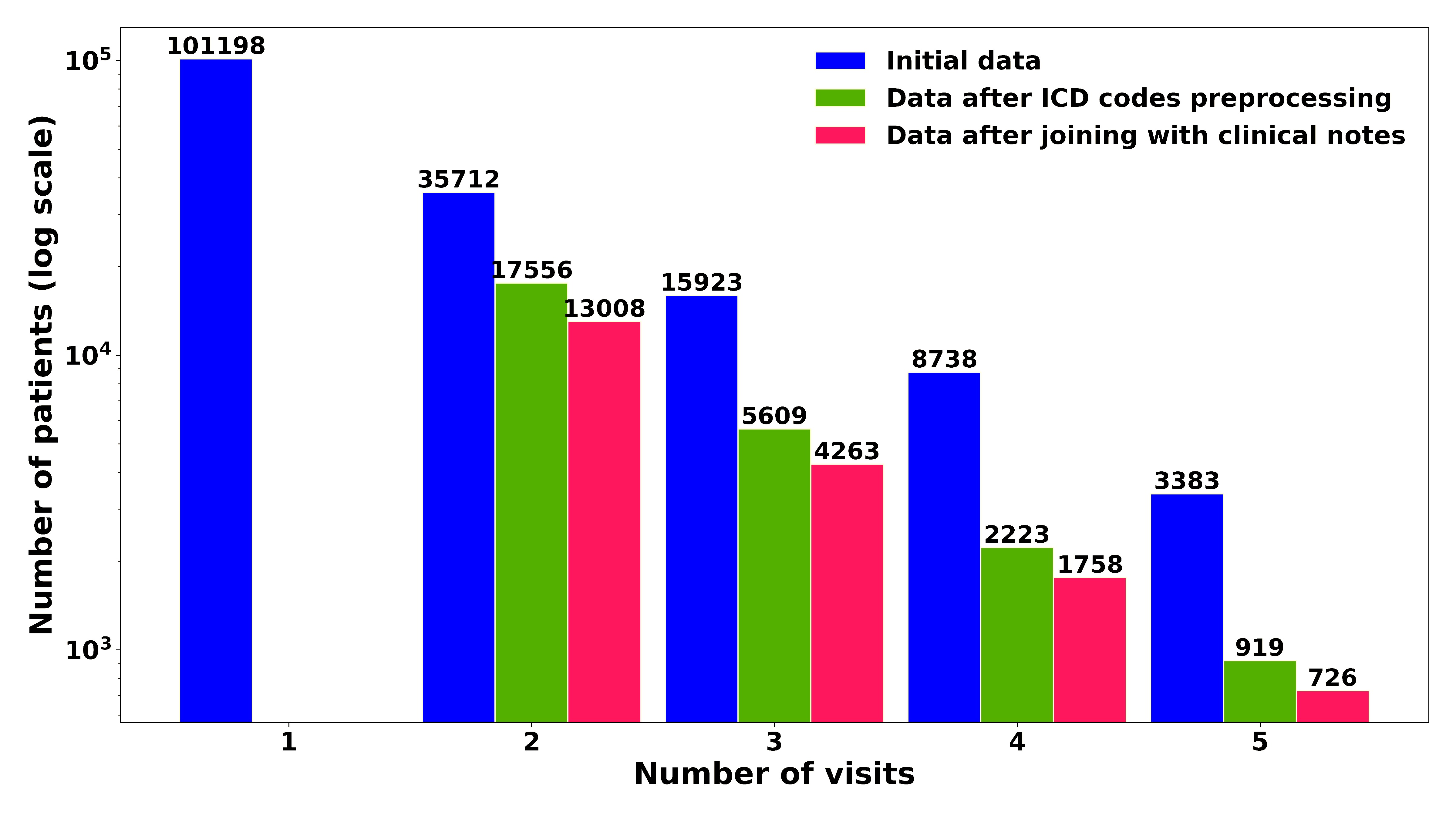}
\caption{Sample distribution of patients by visit count}
\label{fig:distribution}
\end{figure*}

\noindent Table \ref{tab:statistics_of_codes} presents code statistics before and after applying processing steps. This is further illustrated in Figure \ref{fig:distribution}, which reveals a significant data imbalance, showing that a larger number of patients made only a single visit compared to those with multiple visits. 

In addition to structured data processing, clinical notes were also preprocessed to ensure consistency. Working with limited textual datasets poses challenges, particularly due to subword tokenizers that fragment similar tokens differently due to slight structural variations. Therefore, we standardized clinical notes by unifying medical abbreviations (e.g., "hr", "hrs", and "hr(s)" to "hours"), removing accents, converting Danish characters (such as "æ" to "ae"), and putting all notes in lowercase, following the approach of \cite{alsentzer2019publicly}.

\begin{table}[htbp]
\footnotesize
\centering
\begin{tabular}{lccc}
\hline
Code Type & At loading & After preprocessing \\
\hline
\multirow{2}{*}{Procedure codes} & 8482 & 470 \\
 & 3.03 $\pm$ 2.81 & 2.99 $\pm$ 2.77 \\
\hline
\multirow{2}{*}{Diagnosis codes} & 15763 & 762 \\
 & 12.50 $\pm$ 7.67 & 13.18 $\pm$ 8.58 \\
\hline
\multirow{2}{*}{Drug codes} & 1609 & 1609 \\
 & 24.12 $\pm$ 28.19 & 24.12 $\pm$ 28.19 \\
\hline
\end{tabular}
\caption{Code statistics before and after processing}
\label{tab:statistics_of_codes}
\begin{tablenotes}
\small
\item Note: For each code type, the first row shows the number of distinct codes, and the second row shows the mean $\pm$ standard deviation per visit.
\end{tablenotes}
\end{table}

\subsection{Integration of Clinical Notes}
Electronic Health Records (EHRs) typically contain both structured data (e.g., ICD and CCS codes) and unstructured clinical notes. While structured codes provide a standardized representation of diagnoses and procedures, we hypothesize that clinical notes contain additional valuable information that may not be fully captured by these codification systems.

In light of this, we propose incorporating clinical note embeddings into transformer-based models, which have traditionally focused solely on clinical codes. Our model, \textit{Clinical Mosaic}, aims to leverage both structured and unstructured data, offering a more comprehensive view of a patient's clinical state and history. This integration not only enriches the model's input but also enhances its ability to understand and predict patient trajectories more accurately.

\subsubsection{Clinical Mosaic Model}
To effectively exploit the information contained in clinical notes, it is crucial to obtain vector representations. BERT models \cite{devlin2018bert}, and particularly \textit{Clinical BERT} \cite{alsentzer2019publicly}, have proven their ability to capture relevant semantic representations in the medical domain. \textit{Clinical BERT} is a pre-trained model on MIMIC-III, specifically designed for medical notes. However, this model has certain limitations that may affect its performance in our current context:

\begin{enumerate}
    \item \textbf{Limited sequence length:} Clinical BERT was primarily pretrained on sequence lengths of 128 tokens. This limitation may cause the model to under perform when generating representations for longer clinical texts, such as comprehensive discharge summaries. Many studies \cite{wang2024beyond} show that models trained with larger context lengths tend to outperform those trained on shorter sequences, as they can capture more long-range dependencies and contextual information.
    
    \item \textbf{Outdated training data:} \textit{Clinical BERT} was pretrained on MIMIC-III, which is an older version of the MIMIC database. We are currently using MIMIC-IV-NOTES 2.2, which contains more recent and potentially more diverse clinical data. To the best of our knowledge, no publicly available model has been pretrained on this latest version of MIMIC-IV-NOTES. 
\end{enumerate}
These limitations may hinder the model’s ability to fully capture the richness and complexity of clinical narratives. The mismatch between the pretraining data (MIMIC-III) and the target data (MIMIC-IV-NOTES 2.2) could result in suboptimal performance due to differences in language patterns, terminology, and structure.

To address this, we introduce \textit{Clinical Mosaic}, an adaptation of the Mosaic BERT architecture \cite{portes2024mosaicbert} designed for clinical text. The model is pretrained with a sequence length of 512 tokens, leveraging \textit{Attention with Linear Biases} (ALiBi) to improve extrapolation beyond this limit without requiring learned positional embeddings. This allows for better generalization to downstream tasks that may require longer contexts. Pretraining is conducted on 331,794 clinical notes (approximately 170 million tokens) from MIMIC-IV-NOTES 2.2, utilizing 7 A40 GPUs with distributed data parallelism (DDP). The training parameters are detailed in Table \ref{table:clinical_mosaic_params}. To facilitate further research and reproducibility, we publicly release the model weights.\footnote{\url{https://huggingface.co/Sifal/ClinicalMosaic}}

\begin{table}[h]
\centering
\small
\begin{tabularx}{\linewidth}{|l|X|} 
\hline
\textbf{Parameter} & \textbf{Value} \\
\hline
Effective Batch Size & 224 \\
\hline
Training Steps & 80,000 \\
\hline
Sequence Length & 512 tokens \\
\hline
Optimizer & AdamW \\
\hline
Initial Learning Rate & 5e-4 \\
\hline
Learning Rate Schedule & Linear warmup for 33,000 steps, then cosine annealing for 46,000 steps \\
\hline
Final Learning Rate & 1e-5 \\
\hline
Masking Probability & 30\% \\
\hline
\end{tabularx}
\caption{Training parameters of the Clinical Mosaic model}
\label{table:clinical_mosaic_params}
\end{table}

\subsubsection{Evaluation of Clinical Reasoning of Clinical Mosaic}

To evaluate the performance of Clinical Mosaic, we fine-tuned the model on the \textit{Medical Natural Language Inference} (MedNLI) dataset \cite{romanov-shivade-2018-lessons}. MedNLI is a dataset designed for natural language inference tasks in the clinical domain, derived from MIMIC-III clinical notes. It consists of 14,049 pairs of premises and hypotheses, with the objective of classifying the relationship between each pair as entailment, contradiction, or neutral. We report our results on the same test set used by other models for consistency and comparability.

The MedNLI task evaluates several essential aspects of clinical language understanding, including semantic comprehension of medical terminology, logical reasoning in a clinical context, as well as the ability to discern nuanced relationships between clinical statements. Performance on this dataset serves as an indicator of a model's ability to understand and reason about clinical language, a crucial foundation for predicting patient trajectories.

\paragraph{Experimental Setup} 
We fine-tuned Clinical Mosaic using the AdamW optimizer with a linear warmup and decay learning rate schedule. The backbone of the model was initialized from the publicly released \texttt{Sifal/ClinicalMosaic} checkpoint on Hugging Face. The classifier head was trained with an increased learning rate compared to the backbone, allowing for targeted adaptation to the MedNLI classification task. The batch size was set to 64, with gradient clipping applied to stabilize training. We used early stopping with a patience of 10 epochs based on validation loss. The model achieved its best result at epoch 27, with an average loss of 0.0221 and a validation accuracy of 86.5\%.

From the few experiments we conducted, we found the model to be highly sensitive to the learning rate. This could be an artifact of using a lower Adam $\beta_2$ (0.98) than the commonly used value of 0.999, potentially affecting the optimizer's adaptation dynamics.

\begin{table}[h!]
\small
\centering
\begin{tabular}{|c|c|}
\hline
\textbf{Hyperparameter} & \textbf{Value} \\ \hline
Learning Rate (Backbone) &  2e-5 \\ 
Learning Rate (Classifier) &  2e-4 \\ 
Weight Decay & 1e-6 \\ 
Optimizer & AdamW \\ 
Betas & (0.9, 0.98) \\ 
Epsilon ($\epsilon$) & 1e-6 \\ 
Batch Size & 64 \\ 
Max Gradient Norm & 1.0 \\ 
Warmup Epochs & 5 \\ 
Total Epochs & 40 \\ 
Early Stopping Patience & 10 \\ 
Freeze Backbone & False \\  \hline
\end{tabular}
\caption{Hyperparameter configuration used for fine-tuning Clinical Mosaic on MedNLI.}
\label{tab:hyperparams}
\end{table}

Table \ref{tab:mednli} presents the performance of \textit{Clinical Mosaic} on the MedNLI task in comparison with other state-of-the-art models, including the original \textit{Clinical BERT} model \cite{alsentzer2019publicly}.

\begin{table}[h!]
\small
\centering
\begin{tabular}{|c|c|}
\hline
\textbf{Model} & \textbf{Accuracy} \\ \hline
BERT & 77.6\% \\ 
BioBERT & 80.8\% \\ 
Discharge Summary BERT & 80.6\% \\ 
Clinical Discharge BERT  & 84.1\% \\ 
Bio+Clinical BERT & 82.7\% \\ 
\textbf{Clinical Mosaic} & \textbf{86.5\%} \\ \hline
\end{tabular}
\caption{Comparison of performance of BERT variants and Clinical Mosaic on the MedNLI test set.}
\label{tab:mednli}
\end{table}

The results show that \textit{Clinical Mosaic} achieves superior accuracy (86.5\%) compared to existing models, including the original Clinical BERT (84.1\%). This improvement suggests that our model optimizations and pre-training approach have strengthened its clinical language comprehension capabilities. To ensure reproducibility, our training script is available in the Github repository.

\subsection{Fusion of Clinical Representations}
\begin{figure*}[h] \centering \includegraphics[width=0.9\linewidth]{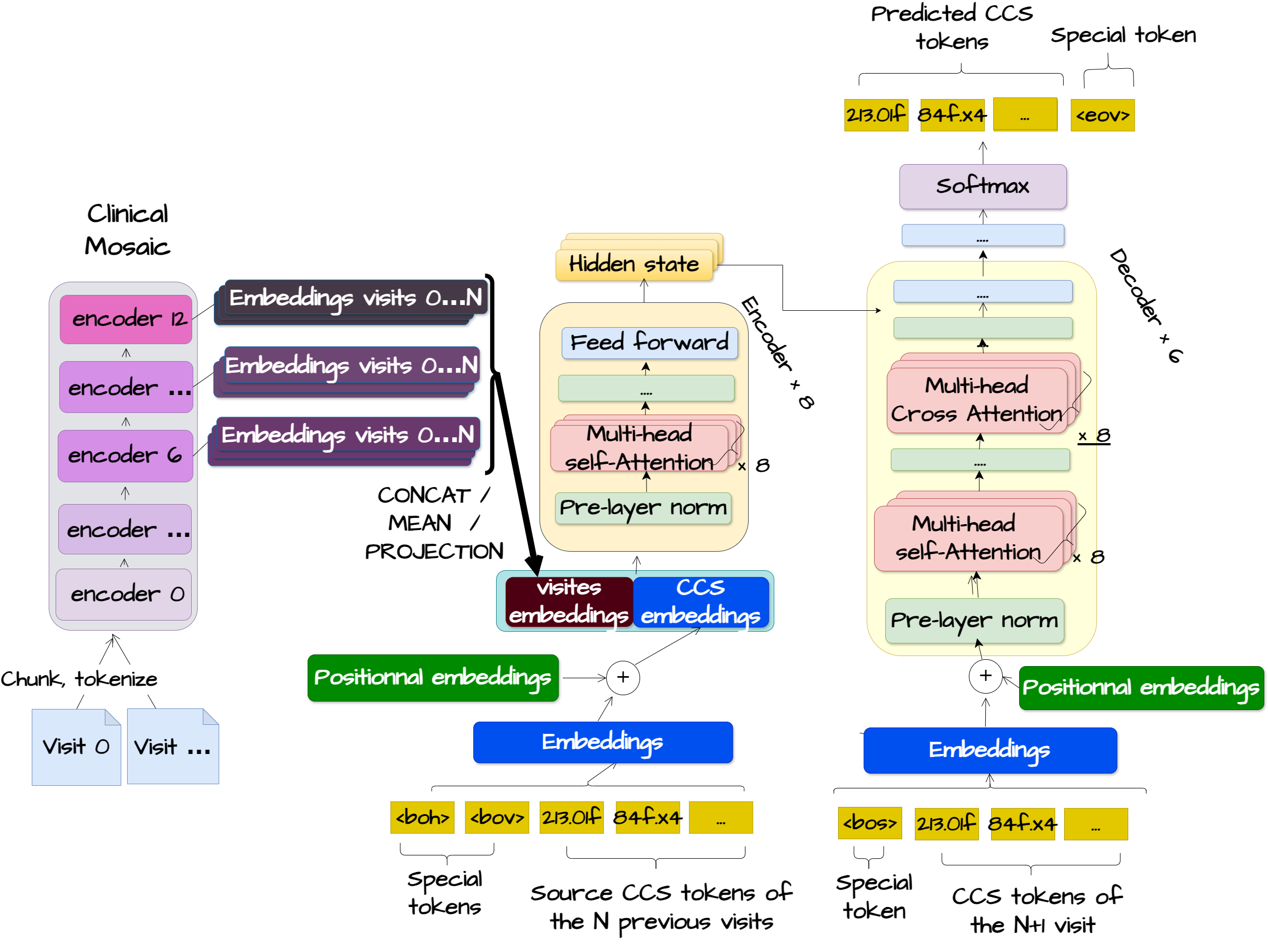} \caption{Architecture for integrating notes} \label{fig:Architecture to integrate the notes} \end{figure*}

To evaluate the impact of integrating clinical note embeddings into an encoder-decoder transformer, we experimented with different fusion points and determined that introducing embeddings at the first layer, before attention mechanisms, led to the best results. This ensures that multi-head attention fully leverages the fused representations, promoting richer interactions (Figure \ref{fig:Architecture to integrate the notes}).

Each layer of BERT encoders generates different representations of clinical notes. Inspired by prior work \cite{hosseini-etal-2023-bert}, which demonstrated the benefits of aggregating multiple layers, we hypothesized that a similar strategy would enhance our clinical tasks. To balance computational efficiency and performance, we aggregated representations from the last six layers of Clincal Mosaic, ensuring robust embeddings while keeping model complexity manageable.

\begin{figure*}[h] \centering \includegraphics[width=0.9\linewidth]{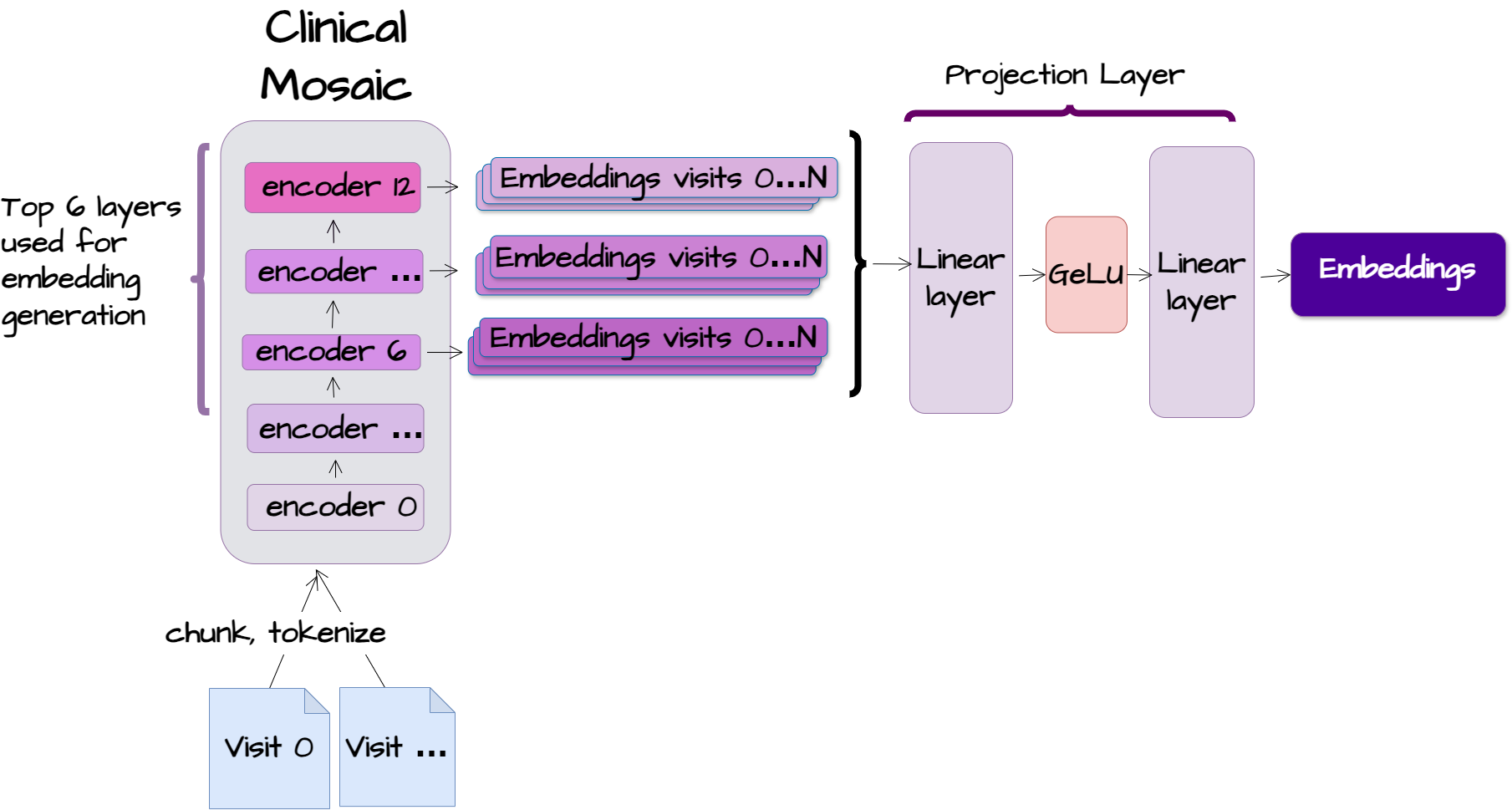} \caption{Approach using a projection layer} \label{fig:Approach of using a projection layer} \end{figure*}

We explored three strategies for embedding generation:

\begin{itemize} \item \textbf{Mean Pooling (MEAN)}: Averages embeddings across six layers and all visits, producing a unified representation that captures overall context while smoothing noise.
\item \textbf{Layer-wise Averaging (CONCAT)}: Averages embeddings only across layers, preserving per-visit representations while reducing dimensionality.
\item \textbf{Projection-based Compression (Projection)}: Projects six-layer embeddings into a lower-dimensional space using a linear layer with GeLU activation. This reduces dimensionality while retaining key information. The projected embeddings are then concatenated to enable the model to learn complex inter-visit relationships (Figure \ref{fig:Approach of using a projection layer}).
\end{itemize}

After generating embeddings, we integrate them with CCS code embeddings within a transformer framework. CCS codes attend to clinical note embeddings through self-attention, forming a unified representation. The decoder then applies causal cross-attention to predict future diagnoses. This fusion of structured (CCS codes) and unstructured (clinical notes) data provides a more comprehensive view of patient trajectories, improving predictive performance.

We also investigated the impact of explicitly adding positional information to the clinical note embeddings, exploring static, sinusoidal, and learned positional embeddings. However, our experiments did not show any noticeable performance improvements. This could be an artifact of the relatively small dataset size, limiting the model's ability to benefit from explicit positional encoding. Alternatively, it is possible that the model implicitly learns to capture temporal relationships through the timestamps included in the clinical notes, making additional positional encoding redundant

\section{Experiments}
\label{sec:results_and_discussion}

In this section, we describe the experiments we conducted to evaluate our approach with the MIMIC-IV and MIMIC-IV-NOTES datasets (37k source, target pairs after preprocessing). The source code is made available for reproducibility purposes\footnote{\url{https://github.com/MostHumble/PatientTrajectoryForecasting}}

\subsection{Metrics}
We evaluate the performance of our models with Mean Average Precision at K (MAP@K) (equation \ref{MapK}) and Mean Average Recall at K (MAR@K) (equation \ref{MarK}) for K=20, 40, 60. These metrics are appropriate for our problem which can be considered as a recommendation task, where order is crucial, and they allow direct comparison with previous work, even if some studies use only one metric \cite{rodrigues2021lig}.

\begin{equation}
\text{MAP@K} = \frac{1}{|Q|} \sum_{u=1}^{|Q|} \frac{1}{\min(m, K)} \sum_{k=1}^K P(k) \cdot rel(k)
\label{MapK}
\end{equation}

\begin{equation}
\text{MAR@K} = \frac{1}{|Q|} \sum_{u=1}^{|Q|} \frac{1}{m} \sum_{k=1}^K rel(k)
\label{MarK}
\end{equation}

Where $|Q|$ is the number of target sequences, $m$ is the number of relevant items in a target sequence, $K$ is the rank limit, $P(k)$ is the precision at rank $k$, and $rel(k)$ is a function that equals 1 if the item at rank $k$ is relevant, 0 otherwise.

\subsection{Baselines}

We compare our approach to state-of-the-art models:

\begin{itemize} 
\item \textit{LIG-Doctor} \cite{rodrigues2021lig}: We use an embedding dimension and a hidden dimension equal to the size of the prediction label, which is 714. This model is followed by a linear layer that merges the bidirectional context with a second linear layer. A softmax layer is then applied. The model is trained for 100 epochs with a patience of 10 epochs (the model converges in 13 epochs), using a batch size of 512 and the Adadelta optimizer.

vbnet

\item \textit{Doctor AI} \cite{choi2016doctor}: We adopted the recommended hyperparameters, with a hidden dimension and embedding dimension set to 2000. A dropout rate of 0.5 is applied. The model is trained over 20 epochs, with a batch size of 384, and uses the Adadelta optimizer.

\item \textit{Clinical GAN} \cite{shankar2023clinical}: The generator uses a 3-layer, 8-head encoder-decoder, with a hidden dimension of 256, and the discriminator uses a 1-layer, 4-head transformer encoder. The model is trained for 100 epochs with a batch size of 8 and converges after 11 epochs. Adam is used for the generator, and SGD for the discriminator, with a Noam scheduler.

\end{itemize}

\subsection{Results} The obtained results are presented in Table \ref{tab:summary}
. All models were evaluated using 5-fold cross-validation and 95\% confidence intervals.

\begin{table*}[htbp]
\centering
\small
\setlength{\tabcolsep}{8pt}
\begin{tabular}{@{}l*{6}{l}@{}}
\toprule
\multirow{2}{*}{Model} & \multicolumn{2}{c}{K = 20} & \multicolumn{2}{c}{K = 40} & \multicolumn{2}{c}{K = 60} \\
\cmidrule(lr){2-3} \cmidrule(lr){4-5} \cmidrule(lr){6-7}
& MAR & MAP & MAR & MAP & MAR & MAP \\
\midrule
Projection & \textcolor{blue}{0.425(5)} & 0.556(21) & \textcolor{blue}{0.439(4)} & 0.556(21) & \textcolor{blue}{0.439(4)} & 0.556(21) \\
Concat & 0.420(6) & \textcolor{blue}{0.569(6)} & 0.425(5) & \textcolor{blue}{0.571(6)} & 0.425(5) & \textcolor{blue}{0.571(6)} \\
Mean & 0.416(6) & 0.538(84) & 0.423(6) & 0.567(17) & 0.423(6) & 0.567(17) \\
Clinical GAN$^1$ & 0.410(5) & 0.558(11) & 0.414(5) & 0.559(12) & 0.414(5) & 0.559(12) \\
Transformer Only & 0.398(23) & 0.565(23) & 0.405(25) & 0.566(23) & 0.405(25) & 0.566(23) \\
LIG-Doctor$^2$ & 0.267(48) & 0.474(94) & 0.361(42) & 0.431(87) & 0.420(37) & 0.402(80) \\
Doctor AI$^3$ & 0.233(5) & 0.206(46) & 0.233(5) & 0.207(47) & 0.233(5) & 0.207(47) \\
\bottomrule
\multicolumn{7}{@{}l@{}}{\footnotesize $^1$\cite{shankar2023clinical}, $^2$\cite{rodrigues2021lig}, $^3$\cite{choi2016doctor}}\\
\multicolumn{7}{@{}l@{}}{\footnotesize Note: Values are presented as mean(standard deviation). For example, 0.425(5) represents 0.425$\pm$0.005.}
\end{tabular}
\caption{Performance of different models using MAP@k and MAR@k. Values are presented as mean(standard deviation in the last decimal place).}
\label{tab:summary}
\end{table*}

The first key observation is that injecting clinical note embeddings into the architecture significantly improves performance, particularly in terms of MAR@K (refer to Figure~\ref{fig average recall @ 20, 40 and 60 for different models}). However, we consider that this improvement could be hindered by the limited size of our dataset (37k samples), preventing the model from fully learning to exploit the representations of the injected embeddings. The strategy of averaging the embedding layers and visits (\textit{Mean}) yields the lowest MAR@K among the embedding injection approaches. This may be due to excessive compression of information, leading to information loss. However, this method is the most computationally efficient, as it only adds one vector. This efficiency is important, particularly due to the $O(N^2)$ computational complexity of the transformer's attention mechanism. 

\begin{figure*}[t] \centering \includegraphics[width=1\linewidth]{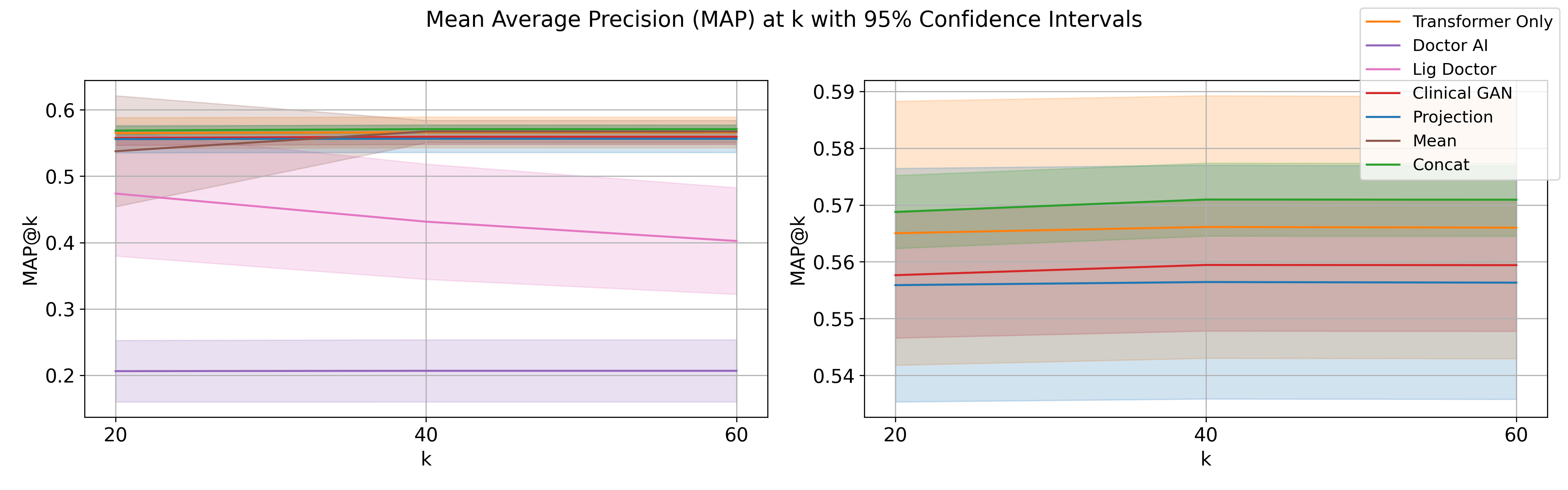} \caption{Mean average precision @ 20, 40, and 60 for different models} 
\label{fig
average precision @ 20, 40 and 60 for different models} 
\end{figure*}

The concatenation method, which averages only the embedding layers, brings significant improvements in terms of MAP@K  as shown in Figure \ref{fig
average precision @ 20, 40 and 60 for different models} and also outperforms all literature methods in terms of MAR@K, with low variance scores across different cross-validation splits. This can be justified by two main reasons: on one hand, the rich representation obtained from the notes, averaged over several layers, enhances the richness of information while preserving critical elements, unlike the \textit{Mean} approach that loses information. On the other hand, this approach also allows the model to be more selective in processing information, leveraging independent elements from different medical visits. \textit{LIG-Doctor} is designed as a classification task, in which a linear layer is used to predict subsequent diagnoses. This setup introduces two important distinctions in the evaluation. First, the model does not generate predictions in a specific order, making the direct calculation of metrics such as MAP@K impossible. To address this issue, we propose sorting the logits to establish a generation order. However, since the model was not trained with this information in mind, performance does not improve efficiently as K increases, as shown in Figure \ref{fig
average precision @ 20, 40 and 60 for different models}. Second, classification prevents repetitive predictions, improving MAR@K results. Figure \ref{fig
average precision @ 20, 40 and 60 for different models} shows that \textit{LIG-Doctor} benefits from increasing K.

\begin{figure*}[t] \centering \includegraphics[width=1\linewidth]{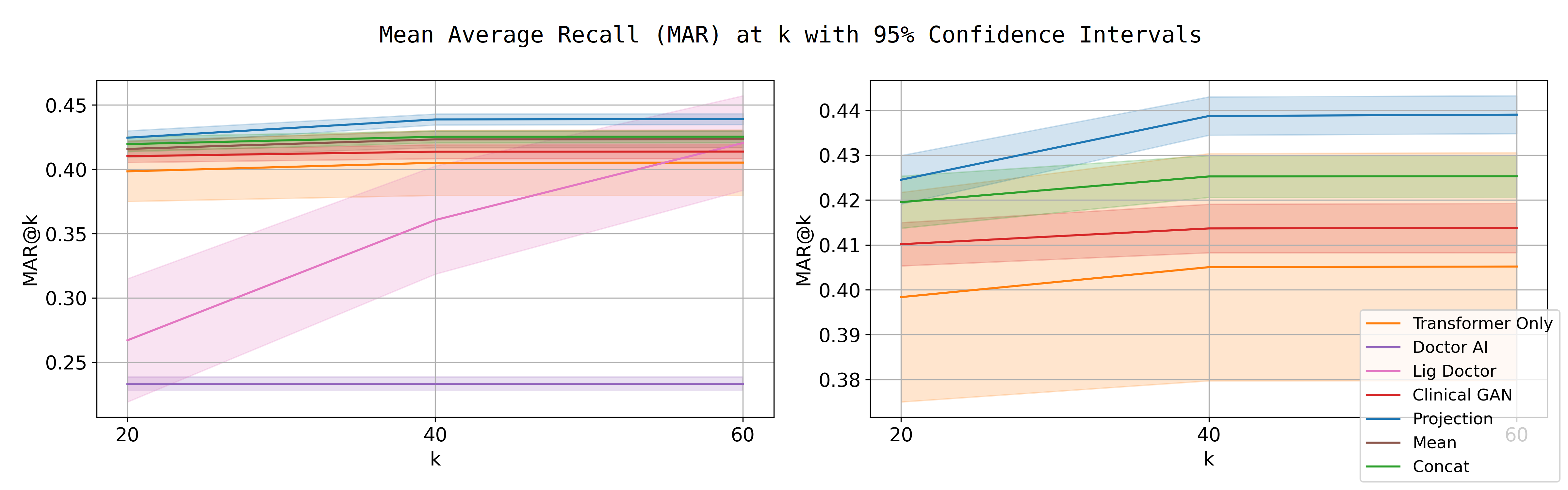} \caption{Mean average recall @ 20, 40, and 60 for different models} \label{fig average recall @ 20, 40 and 60 for different models} \end{figure*}

The results of \textit{Doctor AI} are lower than those of other models, as the model relies on a single GRU layer. Performance appears to be affected by the increase in the prediction space dimension, and improvements could be made by increasing the hidden dimensions and the number of GRU layers. \textit{Clinical GAN} shows good results in terms of MAP@K, but struggles to generate a broader set of relevant predictions, as indicated by its lower MAR@K scores (see Figure~\ref{fig average recall @ 20, 40 and 60 for different models}). This model also exhibits instability during training, a frequent issue with GAN-based architectures, limiting their scalability. This limitation could not be fully explored due to the small size of the dataset used, leaving this question open for future research.

\section{Conclusion} 
\label{sec:Conclusion} 
In this study, we addressed the challenge of predicting patient trajectories by leveraging complementary information from clinical notes to enhance predictive accuracy. Our approach integrates clinical note embeddings into transformer models to forecast patient disease trajectories based on their electronic medical records (EMRs). By combining structured medical data with rich, unstructured information from clinical notes, this method offers a more comprehensive view, potentially leading to more accurate predictions of patient outcomes.

Our experimental results on MIMIC-IV datasets showed that the proposed approach significantly outperforms traditional models that rely solely on structured codes. These findings highlight the considerable potential of utilizing unstructured medical information to improve predictive modeling in healthcare, with the possibility of transforming patient care and resource allocation.

For future work, we plan to investigate strategies for incrementally updating embeddings without disrupting the overall pipeline. A key challenge is ensuring that newly generated embeddings remain compatible with previously learned representations while minimizing computational overhead. We will explore continual learning techniques and efficient adaptation mechanisms to maintain model stability and prevent catastrophic forgetting.  

Additionally, we aim to develop a more automated framework that integrates medical coding with predictive modeling, creating a seamless end-to-end system for clinical decision support. This involves leveraging structured (e.g., CCS codes) and unstructured (e.g., clinical notes) data within a unified architecture, reducing manual intervention in feature extraction and improving interpretability. A fully automated pipeline could enable real-time adaptation to evolving medical knowledge, enhancing predictive accuracy and clinical utility.

\section*{\uppercase{Acknowledgements}}

The project leading to this publication has received funding from the Excellence Initiative of Aix Marseille Université - A*Midex, a French “Investissements d’Avenir programme” AMX-21-IET-017. 

We thank LIS | Laboratoire d'Informatique et Systèmes, Aix-Marseille University for providing the GPU resources necessary for pre-training and conducting extensive experiments. Additionally, we acknowledge CEDRE | CEntre de formation et de soutien aux Données de la REcherche, Programme 2 du projet France 2030 IDeAL for supporting early-stage experiments and hosting part of the computational infrastructure.

 \bibliographystyle{apalike}   
\bibliography{example}

\end{document}